\documentclass[11pt,a4paper]{article}

\usepackage[T1]{fontenc}
\usepackage[utf8]{inputenc}
\usepackage[english]{babel}
\usepackage{amsmath,amssymb}
\usepackage{graphicx}
\usepackage{siunitx}
\usepackage{booktabs}
\usepackage{array}
\usepackage{subfigure}
\usepackage{multirow}
\usepackage{enumitem}
\usepackage{microtype}
\usepackage{cite} 
\usepackage{xcolor}
\usepackage[hidelinks]{hyperref}
\usepackage{cleveref} 
\usepackage{tikz,tikz-3dplot} 
\usetikzlibrary{arrows.meta,angles,decorations.pathmorphing,patterns}

\graphicspath{{./Fig/}}
\newcommand{\NOM}{\textsc{NOM}}
\newcommand{\TLC}{\textsc{TLC}}
\newcommand{\FLC}{\textsc{FLC}}
\newcommand{\NOREF}{\textsc{NO-REF}}
\newcommand{\LT}{\textit{Lap Time}}

\newcommand{\VP}{\textsc{VP}} 
\newcommand{\MM}{\textsc{MM}} 

\makeatletter
\renewcommand*{\@fnsymbol}[1]{\ifcase#1\or 1\or 2\or 3\or 4\or 5\or 6\or 7\or 8\or 9\fi}
\makeatother

\usepackage[top=2.5cm, bottom=2.5cm, left=2.5cm, right=2.5cm]{geometry}

\newcolumntype{L}[1]{>{\raggedright\let\newline\\\arraybackslash\hspace{0pt}}p{#1}}
\newcolumntype{C}[1]{>{\centering\let\newline\\\arraybackslash\hspace{0pt}}p{#1}}
\newcolumntype{R}[1]{>{\raggedleft\let\newline\\\arraybackslash\hspace{0pt}}p{#1}}

\newcommand{\RMS}[1]{\operatorname{RMS}\,(#1)}

\newcommand{\Si}{\Sigma}

\newcommand{\Om}{\Omega}

\newcommand{\be}{\beta}

\newcommand{\bSi}{\boldsymbol{\Si}}
\newcommand{\bOm}{\boldsymbol{\Om}}  
\newcommand{\bxi}{\boldsymbol{\xi}}  
\newcommand{\bmu}{\boldsymbol{\mu}}

\newcommand{\bPsi}{\boldsymbol{\Psi}}
\newcommand{\bPsimu}{\bPsi^{\mu}}%
\newcommand{\bPsiP}{\bPsi^{\text{P}}}%
\newcommand{\dbmu}{\dot{\bmu}}

\newcommand{\bu}{\boldsymbol{u}}

\newcommand{\bw}{\boldsymbol{w}}
\newcommand{\bx}{\boldsymbol{x}}     

\newcommand{\bz}{\boldsymbol{z}}

\newcommand{\bA}{\boldsymbol{A}}     
\newcommand{\bB}{\boldsymbol{B}}

\newcommand{\bI}{\boldsymbol{I}}

\newcommand{\bP}{\boldsymbol{P}}
\newcommand{\bQ}{\boldsymbol{Q}}

\newcommand{\bW}{\boldsymbol{W}}

\newcommand{\bzero}{\boldsymbol{0}}

\newcommand{\dbx}{\dot{\bx}}     

\newcommand{\dbP}{\dot{\bP}}

\newcommand{\calN}{\mathcal{N}}

\newcommand{\calI}{\mathcal{I}}

\DeclareMathOperator{\Prob}{Pr}

\newcommand{\til}{~}

\title{On Disturbance-Aware Minimum-Time Trajectory Planning: Evidence from Tests on a Dynamic Driving Simulator}

\author{Matteo~Masoni\thanks{Università di Pisa, Largo L. Lazzarino 1, 56122 Pisa, Italy.}~,
Vincenzo~Palermo$^1$, Marco~Gabiccini$^1$, Martino~Gulisano$^1$ \\
Giorgio~Previati\thanks{Politecnico di Milano, Via G. La Masa 1, 20156 Milano, Italy.}~,
Massimiliano~Gobbi$^2$, Francesco~Comolli$^2$\\
Gianpiero~Mastinu$^2$, and Massimo~Guiggiani$^1$}

\date{}

\begin{document}

\maketitle

\begin{abstract}
This work aims to investigate how disturbance-aware, robustness-embedding reference trajectories translate into actual driving performance when executed by professional drivers in a dynamic driving simulator. The study compares three planned reference trajectories against a free-driving baseline (\NOREF) to assess the trade-offs between lap time performance and steering effort: \NOM, the nominal time-optimal trajectory; \TLC, a track-limit-robust, time-optimal trajectory obtained by tightening margins to the track edges; and \FLC, a friction-limit-robust, time-optimal trajectory obtained by tightening against axle/tire saturation. All reference trajectories share the same minimum lap-time objective with a small steering-smoothness regularizer, and are evaluated with two professional drivers driving a high-performance car on a virtual track.

The reference trajectories stem from a \emph{disturbance-aware} minimum-lap-time framework recently proposed by some of the authors, where worst-case disturbance growth is propagated over a finite horizon and used to \emph{tighten} tire-friction and track-limit constraints, preserving performance while delivering probabilistic safety margins.

Lap time (LT) and steering energy (SE) are evaluated as indicators of driving performance and steering effort, respectively, while RMS values of lateral deviation, speed error, and drift angle are used to characterize driving style. The results reveal a Pareto-like trade-off between LT and SE: NOM   achieves the shortest LT but with the highest SE, TLC  minimizes SE at the expense of longer LT, while FLC  lies near the efficient frontier, markedly reducing SE relative to NOM with only a minor LT increase. Removing reference trajectories (NOREF) leads to both higher SE and longer LT, confirming that trajectory guidance improves pace and control efficiency. Overall, the findings highlight reference-based and disturbance-aware planning, particularly the FLC  variant, as effective tools for training and for achieving fast yet stable trajectories.
\end{abstract}


\section{Introduction}

Vehicle stability is a central topic in automotive engineering, as it concerns a road vehicle’s ability to maintain a desired trajectory and attitude when subjected to external disturbances, driver inputs, and variable road conditions. From a dynamics perspective, stability encompasses the control of lateral, longitudinal, and vertical motions, and is strongly influenced by vehicle parameters such as mass distribution, suspension geometry, tire characteristics, and steering system design.

Over the last decades, the growing demand for safety and performance has motivated extensive research on the mechanisms leading to loss of stability, including skidding, spin-out, and rollover, as well as on the development of active control systems to prevent them. Advanced systems such as Anti-lock Braking Systems (ABS), Traction Control Systems (TCS), and Electronic Stability Control (ESC) exploit real-time sensing and actuation to compensate for the intrinsic limitations of the vehicle-road-driver system, significantly improving handling robustness in critical maneuvers. In this context, both experimental testing and high-fidelity simulation play a key role in understanding stability limits and in validating new control strategies under reproducible and safety-critical scenarios.

Several strategies have been proposed in the scientific literature to assess vehicle stability. A common practice in control-oriented studies is to impose empirical bounds on selected state variables. Limits on the vehicle sideslip angle, for instance, are frequently employed in yaw-stability and ESC design \cite{Chen:ESC:2014,Lenzo:YawRateSideslip:2021}.
A more model-based research line analyzes simplified lateral-yaw dynamics---typically with single-track formulations---to characterize equilibrium configurations and their stability through bifurcation methods. Comprehensive continuation studies have revealed phenomena such as saddle-node and Hopf transitions in response to changing inputs and tire characteristics \cite{DellaRossa:bifurcation:2012}.

Beyond the identification of stable steady states, an important question in nonlinear dynamics concerns the estimation of the Regions of Attraction (ROAs), i.e., the sets of initial conditions that converge asymptotically to a desired equilibrium. Recent work has estimated ROAs by adopting enhanced tire models \cite{Hashemi:stability:2016} or multi-point linearization techniques \cite{Huang:StabilityRegionsVehicle:2020}. Another promising approach is based on Sum-of-Squares (SOS) programming, which enables systematic ROA computation through polynomial Lyapunov functions obtained via semidefinite programming. Several applications to vehicle lateral dynamics have already used SOS techniques to estimate stability regions \cite{Ribeiro:SOSapproach:2020,Zhu:SOSBasedVehicle:2022}. 

Alternative quantitative indicators include Lyapunov exponents to measure convergence rates to capture how disturbances decay during aggressive maneuvers \cite{Sadri:LyapunovExponents:2013}. However, aymptotic stability indicators offer limited operational guidance for planning: a point along the planned trajectory may lie inside the ROA of a stable equilibrium even if that equilibrium is reached far ahead in time or outside the admissible domain (e.g., off track), which makes asymptotic guarantees only loosely connected to the short-horizon constraint satisfaction required in practice.

Building on theoretical studies that analyze the mechanisms of vehicle stability and loss of control, dynamic driving simulators offer a powerful experimental platform to test and further investigate these phenomena in a controlled, repeatable and risk-free environment. By allowing critical maneuvers and near-limit conditions to be safely reproduced, simulators enable systematic assessment of how drivers and vehicles behave close to the stability boundaries identified by theory, and how control strategies perform under such conditions. At the same time, their usefulness depends on how well simulator behavior reproduces real driving: a recent systematic review \cite{Zhang:DrivingSimulatorValidation:2025} shows that simulators can often achieve acceptable validity, but also highlights cases of mismatch and methodological limitations, underlining the need for rigorous validation and transparent reporting when using simulators for vehicle stability studies.

Driving simulators have been widely used to investigate vehicle stability in specific safety-related contexts. In \cite{Maruyama:DrivingSimulatorExperiment:2006}, simulator tests with different vehicle speeds and crosswind profiles, including uniform and transient gusts, are carried out to study how crosswind-induced forces affect lateral deviation, yaw response and controllability on expressways. The resulting data are used to support safety-related decisions, such as defining expressway closure criteria and designing wind barriers.
Driving simulators have also been employed to study sudden vehicle faults and their impact on controllability. In \cite{Wanner:SingleWheelHub:2016}, drivers perform motorway-speed simulator tests in which single wheel hub motor failures are artificially triggered under straight-line and cornering conditions, creating strong yaw disturbances and longitudinal decelerations. The objective of these experiments is to quantify driver reaction times and resulting path deviations, and to assess the safety-criticality of such failures.
Driving-simulator experiments on truck platooning are conducted in \cite{Cho:DevelopmentStabilityIndex:2025} to develop and validate a quantitative stability index for drivers' psychological stability under different time gaps and visualization conditions (with/without a see-through function). The tests relate vehicle-motion metrics and visual information to perceived comfort and safety in closely spaced platoons, addressing driver acceptance and subjective stability.

Beyond safety-oriented applications, driving simulators have been instrumental in validating theoretical frameworks for vehicle stability and loss of control. A further line of research focuses on the straight-ahead stability of passenger cars. In \cite{Mastinu:StraightRunningStability:2020}, nonlinear bifurcation analysis is combined with experiments in a dynamic driving simulator to study how passenger cars lose straight-running stability as speed and disturbance levels increase, for both understeering and oversteering setups. In this case, the simulator experiments primarily serve to validate theoretical predictions about stability boundaries and loss-of-control mechanisms in straight-line driving.
A broader theoretical perspective on stability with the driver in the loop is provided in \cite{Mastinu:GlobalStabilityRoad:2023}, where a nonlinear vehicle-driver model is analyzed using bifurcation theory and Lyapunov methods, and then validated through experiments in a dynamic driving simulator. In this study, simulator runs are mainly used to confirm and illustrate the predicted domains of attraction and loss-of-control trajectories after disturbances, embedding them into a global stability framework for everyday road driving.
In \cite{Mastinu:HowDriversLose:2024}, dynamic models of the vehicle-driver system are combined with driving-simulator and track experiments to analyze how drivers lose control after strong disturbances such as severe lane changes, wind gusts or road irregularities. The test program concentrates on the onset and evolution of instability and loss-of-control trajectories near the stability boundary.

Understanding driver behavior and interaction with the vehicle is crucial for comprehensive stability analysis. In \cite{Previati:SAEIntJVehDynStabNVH:2024}, the authors investigate, at a dynamic driving simulator, how the driver interacts with the steering wheel during cornering, in order to detect driving strategies. Such driving strategies allow to derive accurate holistic driver models for enhancing safety.
Another important application of driving simulators is comparative vehicle dynamics testing. In \cite{Kharrazi:VehicleDynamicsTesting:2020}, motion-based simulator experiments with heavy vehicles are used to evaluate whether drivers can reliably perceive handling differences induced by systematic variations of key parameters such as suspension and steering characteristics, combining subjective ratings with objective measurements. The main goal of these tests is to support early-stage vehicle dynamics development and parameter tuning in a cost-effective way.
The handling performance of ultra-efficient lightweight vehicles (quadricycles) is studied in a dynamic driving simulator in \cite{Musso:SAEIntJAdvCurrPracMobility:2024}. The proper mechanical actions on the driver make the simulations trustworthy, enabling reliable assessment of handling characteristics for this class of vehicles.

Moving toward control-oriented applications, driving simulators enable validation of active systems designed to compensate for disturbances and enhance stability. In \cite{Asperti:SAEIntJVehDynStabNVH:2025}, driver-in-the-loop simulations at a dynamic driving simulator were performed for deriving a new electric power steering system able to give proper steering feedback in case of high torque vectoring at the front axle. Such a torque can be seen as a disturbance to the driver, and the simulator tests enable validation of the compensation strategy under realistic driving conditions.
A more control-oriented use of driving simulators is presented in \cite{Alfatti:ImplementationPerformancesEvaluation:2023}, where a real-time Hardware-in-the-Loop setup is used to compare advanced lateral stability controllers (LQR and sliding-mode) against a commercial ESP through standardized but stressed maneuvers such as step steer, sine steer and lane-change. These tests mainly serve to benchmark and refine stability control strategies near the limits of adhesion in demanding yet repeatable scenarios.
In \cite{Novi:SAEIntJVehDynStabNVH:2018}, future vehicles are studied aiming to exclude the human factor in limiting handling maneuvers. A dynamic driving simulator is used to compare robotic controllers with human drivers, showing that robotic controllers can outperform human drivers and posing interesting design challenges for future automated vehicles.

While most simulator-based studies focus on road-vehicle applications, recent work has begun exploring motorsport contexts. Exploiting a dynamic driving simulator, in \cite{DellaRossa:SAEIntJVehDynStabNVH:2025} the authors show how to detect as quickly as possible whether the driver will lose control of a vehicle, after a disturbance has occurred. A degree of stability for the vehicle-and-driver combination is proposed which seems already applicable for motorsport applications.

Building on these contributions, while \cite{DellaRossa:SAEIntJVehDynStabNVH:2025} touches on motorsport applications, most of the cited simulator-based studies are framed around road-vehicle applications, such as crosswind safety, driveline failures, global stability in everyday driving, platooning comfort, or early-stage vehicle dynamics development, and do not explicitly target sustained, near-limit motorsport operation with robust planning strategies. In contrast, the present work takes the disturbance-aware minimum-lap-time planner recently proposed in \cite{Gulisano:DisturbanceawareMinimumtimePlanning:2025} and brings it to the DriSMi driving simulator\til \cite{drismi:website:2024}, using professional drivers capable of operating at high performance on a race-like track to assess how robustness-embedding criteria translate into actual driving. Specifically, the open-loop, covariance-based framework in~\cite[Sec.3]{Gulisano:DisturbanceawareMinimumtimePlanning:2025} is used to generate three reference trajectories -- nominal time-optimal (\NOM), track-limit-robust (\TLC) and friction-limit-robust (\FLC) -- which are compared with a free-driving baseline (\NOREF). This enables quantification of how robustness-oriented constraint tightening influences lap time, steering effort and tracking/style indicators once human drivers attempt to execute these trajectories at high performance.

The rest of the paper is organized as follows.
Section~\ref{sec:ol_robust_planning} revisits the open-loop disturbance-aware planner, detailing the vehicle model, uncertainty propagation, and the constraint-tightening mechanisms that yield the three reference trajectories (\NOM, \TLC, \FLC).
Section~\ref{sec:vehicle_model_identification} documents how telemetry from the high-fidelity 14-DoF simulator was used to calibrate the single-track surrogate so that planning and execution remain consistent.
Section~\ref{sec:experimental_setting} introduces the drivers, simulator setup, reference-generation protocol, and evaluation metrics, and outlines the randomized test campaign including the free-driving baseline \NOREF.
In Section~\ref{sec:results} lap-time, steer-energy, and tracking indicators are reported and discussed, highlighting the LT--SE trade-offs that emerge from the different references.
Finally, Section~\ref{sec:conclusion} summarizes the main findings and discusses limitations and future research directions.

\section{Open-loop disturbance-aware planning}
\label{sec:ol_robust_planning}

This section recalls the essential notions of the disturbance-aware, robustness-embedding framework employed for trajectory planning. For a comprehensive treatment of the theoretical foundations, implementation details, and algorithmic aspects, the reader is referred to \cite{Gulisano:DisturbanceawareMinimumtimePlanning:2025}.

\providecommand{\bmu}{\boldsymbol{\mu}}
\providecommand{\bx}{\mathbf{x}}
\providecommand{\bu}{\mathbf{u}}
\providecommand{\bz}{\mathbf{z}}
\providecommand{\bxi}{\boldsymbol{\xi}}
\providecommand{\bA}{\mathbf{A}}
\providecommand{\bB}{\mathbf{B}}
\providecommand{\bP}{\mathbf{P}}
\providecommand{\bW}{\mathbf{W}}
\providecommand{\bI}{\mathbf{I}}
\providecommand{\bzero}{\mathbf{0}}
\providecommand{\bOm}{\boldsymbol{\Omega}}
\providecommand{\calI}{\mathcal{I}}
\providecommand{\NOM}{\textsc{NOM}}
\providecommand{\TLC}{\textsc{TLC}}
\providecommand{\FLC}{\textsc{FLC}}
\providecommand{\NOREF}{\textsc{NO-REF}}


\subsection{Vehicle and disturbance model}

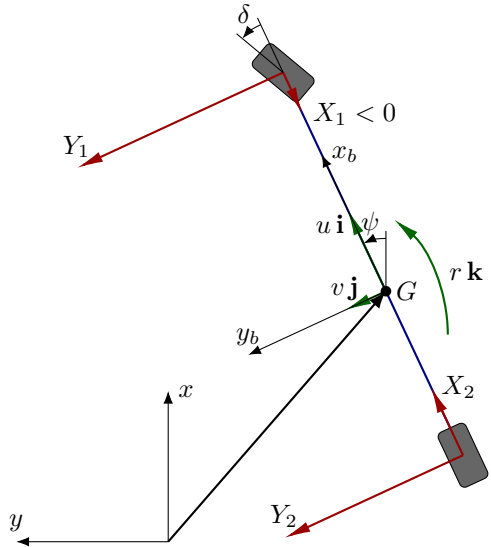
\begin{figure}
	\centering
	\begin{tikzpicture}[scale=2, x = {(0,1cm)}, y = {(-1cm,0)},
	force/.style={->, thick, red!60!black, >={Latex[length=10pt, width=4pt]}},
	vec/.style={->, thick, mgreen, >={Latex[length=10pt, width=4pt]}},
	aux/.style={thin},
	frame/.style={->,thin,>={Latex}},
	every node/.style={font=\small, text=black}
	]
	\definecolor{mgreen}{RGB}{0,100,0}
	\begin{scope}[rotate = 25,shift = {(-.3,-2)}]
		\coordinate (rear) at (0,0);
		\coordinate (G) at (1.2,0);
		\coordinate (front) at (2.8,0);
		
		\draw[blue!50!black, thick] (rear) -- (front);
		
		\draw[fill=black!60!white, draw=black, rounded corners=2pt]
		($(rear)+(-.2,.1)$) rectangle ($(rear)+(+.2,-.1)$);
		
		\begin{scope}[shift={(front)}, rotate=25]
			\draw[fill=black!60!white, draw=black, rounded corners=2pt]
			(-0.2,.1) rectangle (0.2,-.1);
		\end{scope}
		
		\draw[force] (rear) -- ++(.5,0) node[above,right] {$X_2$};
		\draw[force] (rear) -- ++(0,1.3) node[above] {$Y_2$};
		
		\draw[force] (front) -- ++(-.3,0) node[right] {$X_1<0$};
		\draw[force] (front) -- ++(0,1.5) node[above] {$Y_1$};
		
		\draw[vec] (G) -- ++(.6,0) node[left, shift={(-.2,-.1)}] {$u\,\mathbf{i}$};
		\draw[vec] (G) -- ++(0,.3) node[above] {$v\,\mathbf{j}$};
		
		\begin{scope}[rotate = -115]
			\draw[vec] 
			(-.1,.8) arc[start angle=0, end angle=50, radius=1]
			node[midway, right=2pt] {$r\,\mathbf{k}$};
		\end{scope}
		
		\fill (G) circle (1pt) node[right] {$G$};
		
		\draw[frame] (G) --++ (1,0) node[right] {$x_b$};
		\draw[frame] (G) --++ (0,1) node[above] {$y_b$};
		
		
		\draw[aux] (front) -- ++(.4,0);
		\draw[aux] (front) -- ++({0.4*cos(25)}, {0.4*sin(25)});
		
		\draw[->,thin,>={Latex}] ([shift={(0.35,0)}]front) arc[start angle=0, end angle=25, radius=0.35];
		\node[left,shift={(.15,0)}] at ([shift={(0.35,0)}]front) {$\delta$};
		
		\draw[aux] (G) -- ++({0.4*cos(25)}, {-0.4*sin(25)});
		
		\draw[->,thin,>={Latex}] ([shift={({0.35*cos(25)},{-.35*sin(25)})}]G) arc[start angle=-25, end angle=0, radius=0.35];
		\node[left,shift={(.2,-.05)}] at ([shift={({0.35*cos(25)},{-.35*sin(25)})}]G) {$\psi$};
		
	\end{scope}
	\coordinate (ground) at (0,0);
	\draw[frame] (ground) --++ (1,0) node[right] {$x$};
	\draw[frame] (ground) --++ (0,1) node[above] {$y$};
	\draw[->,thick,>=Latex] (ground) --++ (G);
	
\end{tikzpicture}
	\caption{Single-track model used as a basis for our stochastic vehicle dynamics model.}
	\label{fig:vehicle_model}
\end{figure}

The planner relies on the nonlinear Single-Track model\til \cite{Guiggiani:ScienceVehicleDynamics:2023} with nonlinear tire characteristics and open differential schematically depicted in Figure~\ref{fig:vehicle_model}.
The state vector \(\bx = (u,v,r,x_G,y_G,\psi)^{\mathsf T}\) collects longitudinal and lateral velocities, yaw rate, and the pose of the vehicle with respect to the center of mass.
The control input \(\bu = (X_{2,a},X_{2,b},\delta)^{\mathsf T}\) includes the rear-axle acceleration force \(X_{2,a}\), the braking force \(X_{2,b}\) (distributed between axles via brake balance), and the steering angle \(\delta\).
Let \(f(\bx,\bu)\) denote the resulting dynamics, where lateral forces \(Y_j(\bx,\bu)\) follow Pacejka's Magic Formula and vertical loads \(Z_j(\bx,\bu)\) capture load transfer.
In open loop, stochastic perturbations related to the uncertainty on the vehicle state are modelled as additive Gaussian noise \(\bw(t)\),  
\begin{equation}
  \dbx(t) = f\!\big(\bx(t),\bu(t)\big) + \bw(t),\qquad \bw(t)\sim\mathcal{N}(\bzero,\bQ(t)),
  \label{eq:stochastic_vehicle}
\end{equation}
where $\calN(\bzero, \bQ(t))$ denotes a Gaussian distribution with $\bzero$ mean and covariance matrix $\bQ(t)$.
\subsection{Mean and covariance propagation}
Assuming a first-order disturbance propagation, the state becomes Gaussian with mean \(\bmu(t)\) and covariance \(\bP(t)\).
The mean trajectory satisfies
\begin{equation}
  \dbmu(t) = f\!\big(\bmu(t),\bu(t)\big), \quad \bmu(0)=\bmu_0,
  \label{eq:mean_dynamics}
\end{equation}
while the covariance evolves through the Lyapunov differential equation
\begin{equation}
  \dbP(t) = \bA(t)\,\bP(t) + \bP(t)\,\bA^{\mathsf T}(t) + \bQ(t), \quad \bP(0)=\bP_0=\bP_0^{\mathsf T},
  \label{eq:lyapunov_continuous}
\end{equation}
with Jacobian \(\bA(t)=\partial f/\partial \bx|_{(\bmu(t),\bu(t))}\).
This approximation preserves tractability and provides closed-form expressions for the gradients used in the chance-constraint back-offs described in Sec.~\ref{sec:constraint_back-offs}. A more detailed treatment is provided in~\cite{Gulisano:DisturbanceawareMinimumtimePlanning:2025}.

\subsection{Collocation discretization and \(H\)-step predictions}
The track centerline is parameterized by the curvilinear variable \(\alpha\in[0,1]\) and sampled at \(N+1\) nodes \(\alpha_0,\dots,\alpha_N\).
At each interval \([\alpha_k,\alpha_{k+1}]\) we approximate \(\bmu(\cdot)\) and \(\bP(\cdot)\) with degree-\(d\) polynomials \(\pi_k(\tau)\) defined on the normalized interval \(\tau\in[0,1]\) and collocated at Gauss--Legendre points \(\tau_{\ell}\).
The associated collocation states and covariances are denoted \(\bxi_{k,\ell}\) and \(\bSi_{k,\ell}\).
To emulate the absence of closed-loop correction (i.e., the delay in driver corrections), we propagate \(H+1\) covariance replicas:
\(\bP_k^0\) is re-initialized at each node with \(\bar{\bP}_0\succeq 0\),
whereas \(\bP_k^j\) for \(j\geq 1\) captures the covariance evolved \(j\) steps earlier.
The residual maps \(\bPsimu_k(\cdot)\) and \(\bPsiP_k(\cdot)\) enforce, respectively, the \emph{collocation constraints} for \((\bmu,\bxi)\) and the \emph{continuity} of every replica \(\bP_k^j\) across the grid.
Algebraic quantities such as tire forces or track offsets are grouped in \(\bz_k\) and constrained through \(\bOm_k(\cdot)\).

\subsection{Open-loop disturbance-aware OCP}
The disturbance-aware reference generator solves the nonlinear program
\begin{subequations}\label{eq:DOCPopenloop}
\begin{alignat}{3}
  \underset{\bmu_k,\bxi_k, \bu_k, \bP_k,\bz_k}{\text{minimize}} \,
  & & & J_k(\bmu_k,\bxi_k, \bu_k) & & \label{eq:DOCPcostopenloop} \\[2pt]
  \hspace*{-2.0 cm}\text{s.t.} \quad
  & \bzero      & = & \; \bPsimu_k(\bmu_{k-1},\bmu_k,\bxi_k, \bu_k,\bz_k),
  & \quad & k = 1,\ldots, N \label{eq:DOCPdynopenloop} \\
  & \bmu_0      & = & \; \bar{\bmu}_0
  & & \label{eq:DOCPdynICopenloop} \\
  & \bzero      & = & \; \bPsiP_k(\bmu_k,\bxi_k, \bu_k, \bP_{k-1}^{j-1},\bP_k^j,\bSi_k^j,\bz_k),
  & \quad & \substack{k = 1,\ldots, N; \\ j=1,\ldots,H} \label{eq:DOCPdPopenloop} \\
  & \bP_k^0     & = & \; \bar{\bP}_0 \succeq 0
  & \quad & k = 1,\ldots, N \label{eq:DOCPdPinitopenloop} \\
  & \bzero      & = & \; \bOm_k(\bmu_k,\bxi_k, \bu_k,\bz_k),
  & \quad & k = 0,\ldots, N \label{eq:DOCPpathopenloop} \\
  & 0           & \geq & \; h_i(\bmu_k, \bu_k, \bz_k) + \be_i(\bmu_k, \bu_k, \bP_k^H, \bz_k),
  & \quad & k = 1,\ldots, N;\; i \in \calI \label{eq:DOCPconstraintsopenloop}
\end{alignat}
\end{subequations}
where \(J_k\) is the standard minimum-lap-time cost (i.e. our planning objective).
Equations~\eqref{eq:DOCPdynopenloop}--\eqref{eq:DOCPpathopenloop} collect the collocation residuals, initial conditions, and algebraic constraints.
Constraint~\eqref{eq:DOCPconstraintsopenloop} evaluates the safety back-off using the most propagated covariance \(\bP_k^H\), hence embedding worst-case open-loop uncertainty growth over the horizon \(H\). A more detailed description of the constraints formulation and the associated back-off terms is provided in Sec.~\ref{sec:constraint_back-offs}.

\subsection{Constraint back-offs and planner variants}
\label{sec:constraint_back-offs}
For each inequality \(h_i(\bx,\bu,\bz)\le 0\), the deterministic back-off \(\be_i\) is computed from the linearized chance constraint
\(\Prob\{h_i(\bx,\bu,\bz)\le 0\}\ge p_i\) as
\begin{equation}
  \be_i = \Phi^{-1}(p_i)\,\sigma_i,\qquad
  \sigma_i = \big[\nabla_{\bx} h_i(\bmu_k,\bu_k,\bz_k)^{\mathsf T}\,\bP_k^H\,\nabla_{\bx} h_i(\bmu_k,\bu_k,\bz_k)\big]^{1/2},
  \label{eq:backoff}
\end{equation}
with \(\Phi^{-1}\) the standard-normal quantile.

Track boundaries (TLC) are enforced by shifting the allowable lateral offset \(e\) as \(e_{\min}+ \be^{\mathrm{TLC}}\le e \le e_{\max}-\be^{\mathrm{TLC}}\), where \(\nabla_{\bx}h^{\mathrm{TLC}}\) depends on the track normal.

Friction limits (FLC) rely on the axle saturation ratio
\begin{equation}
  S_j(\bx,\bu) = \frac{\left(\frac{X_j(\bx,\bu)}{\mu_{x,j}}\right)^2+\left(\frac{Y_j(\bx,\bu)}{\mu_{y,j}}\right)^2}{Z_j^2(\bx,\bu)}, \qquad j\in\{1,2\},
  \label{eq:axle_saturation_summary}
\end{equation}
constrained by \(S_j - 1 + \be^{\mathrm{FLC}}_j \le 0\).
Choosing \(\be_i\equiv 0\) yields the nominal planner (\NOM). Emphasizing the track limit (\(\be^{\mathrm{TLC}}>0\)) produces the \TLC{} references, while focusing on the friction limit (\(\be^{\mathrm{FLC}}>0\)) produces the \FLC{} trajectories discussed in Sec.~\ref{sec:results}. A thorough theoretical treatment can be found in our previous work~\cite{Gulisano:DisturbanceawareMinimumtimePlanning:2025}.

\emph{Remark.} The horizon \(H\) controls conservatism and computational effort; in practice it is tuned to cover the time span before significant driver corrections intervene.

\subsection{Parameters selection}

The initial covariance $\bP_0$ and the disturbance covariance $\bQ(t)$ are key parameters that govern the magnitude of the back-off terms $\be_i$ 
in~\eqref{eq:backoff}. These terms effectively ``inflate'' the constraints, providing a safety margin that ensures constraint satisfaction under uncertainty. The selection of the above parameters must be based on engineering criteria to balance robustness against excessive conservatism. Specifically, \(\bP_0\) and \(\bQ(t)\) are tuned to ensure that the resulting back-off terms do not become excessively large, which would unnecessarily reduce the potential of the axles by forcing the planner to operate well below the friction limits.
 
Since the effect of these parameters on the back-off terms is not clearly visible a priori, a steering maneuver with high lateral acceleration that saturates the axles above 90\% was simulated. 
During this process, the covariance matrix was propagated for \(H\) steps from each discretization step, and the back-off terms were computed using the most propagated covariance matrix. Based on this analysis, the parameters were tuned via trial and error.

\section{Vehicle model identification procedure}
\label{sec:vehicle_model_identification} 

Planning under uncertainty is computationally demanding: at each time step the covariance matrix is reset and propagated over a horizon of $H$ steps ($H=4$ in our case) to capture the growth of uncertainty in the dynamics. This propagated covariance is then used to determine the amount of constraint tightening required at each $H$-th step. To keep runtime manageable we currently adopted a single-track vehicle model. The driving simulator, however, implements a high-fidelity 14-DoF dynamic model with detailed suspension kinematics; consequently, an identification procedure was required to align the two.

\subsection{DriSMi layout and vehicle dynamics model}
 
The dynamic simulator employed is located at Politecnico di Milano (DriSMi lab~\cite{drismi:website:2024}). It is a new-generation, mid-size facility. A six-actuator electric hexalift provides full six-DoF motion with ~20 Hz closed-loop bandwidth and is mounted on three aerostatic pads that enable frictionless planar travel. Longitudinal and lateral motion on a machined 6 $\times$ 6 m deck is produced by four independent cable drives, which decouple x/y translations and yaw; yaw rotations exceed $60^\circ$, with ~3 Hz bandwidth. The enlarged workspace supports peak accelerations up to 1.5 g (longitudinal/lateral) and 2.5 g (vertical), with maximum end-to-end latency of ~20 ms. To offset moving-mass inertia and protect the cable system, three counter-masses are employed. Immersion is enhanced by five projectors on a $270^{\circ}$ screen, an active seat and belt tensioners for sustained acceleration cues, electric steering torque feedback, an active hydraulic brake (including ABS effects), and a five-speaker audio system. For comfort/NVH studies, eight shakers (bandwidth up to 200 Hz) reproduce engine- and road-induced vibrations.
 
The high-fidelity vehicle representation is a 14-DoF multibody model implemented in the VI-CarRealTime environment~\cite{vigrade:website:2025} (hereafter, the 14-DoF model). The chassis has six degrees of freedom (three translations plus roll, pitch, and yaw). Each of the four wheels contributes two additional DoF (rolling rotation and vertical displacement). Additional implementation details on the 14-DoF model can be found in~\cite{vigrade:website:2025,Mastinu:RoadOffRoadVehicle:2014,Mastinu:StraightRunningStability:2020}.

\subsection{Identification procedure}

The vehicle employed in this study is a high performance car whose main parameters are reported in Table~\ref{tab:singletrack_par}. 
Geometric and inertial parameters and tire characteristics have been measured from the actual vehicle and introduced into the simulator model \cite{Mastinu:StraightRunningStability:2020,Mastinu:GlobalStabilityRoad:2023,Mastinu:HowDriversLose:2024}. Tire forces are modeled via a Pacejka Magic Formula 6.1~\cite{vigrade:website:2025}. These data provided the initial baseline for the single-track model used in planning. The parametrization corresponds to a decidedly oversteer-biased setup-deliberately chosen to make the car more demanding to drive and to accentuate any differences in the designed reference trajectory the driver would follow.  The geometric and inertial parameters employed in the single-track surrogate were extracted directly from the simulator model; the adopted values are summarized in Table~\ref{tab:singletrack_par}.

\begin{table}[h]
	\centering
	\caption{Baseline geometric and inertial parameters extracted from the simulator model and used in the single-track surrogate.}
	\begin{tabular}{ll}
		\hline
		Parameter & Value \\
		\hline
		Mass & $1875$ kg \\
		Yaw inertia & $3341$ kg\,m$^2$ \\
		Wheelbase & $2.97$ m \\
		Weight distribution (front/rear) & $53/47$ \\
		CoG height & $0.5$ m \\
		\hline
	\end{tabular}

	\label{tab:singletrack_par}
\end{table}

However, when driven with identical inputs, the two models produced noticeably different responses at high lateral acceleration levels. The mismatch stems both from neglected body dynamics in the single-track and from the fact that axle characteristics represented with an MF structure are not directly mappable from the per-tire parameters of the 14-DoF simulator model.

To model the lateral forces at the front and rear axles of the single-track vehicle we adopted a simplified version of the pure side slip Magic Formula~\cite{Pacejka:book:2012}. In particular, for each axle we assume
\begin{align}
\label{eq:mf_simplified}
	F_{y} (\alpha,F_{z};\boldsymbol{p}) &= D_y \sin\!\bigl(C_y \arctan(B_y \alpha) - E_y(B_y \alpha - \arctan(B_y \alpha))\bigr),
\end{align}
with $\boldsymbol{p} = (F_{z0}, p_{Cy1}, p_{Dy1}, p_{Dy2}, p_{Ey1}, p_{Ey2}, p_{Ky1}, p_{Ky2})$ and where
\begin{align*}
    df_z &= \frac{F_z - F_{z0}}{F_{z0}}, & D_y &= \bigl(p_{Dy1} + p_{Dy2}\,df_z\bigr) F_z,\\
    C_y &= p_{Cy1},                      & B_y &= p_{Ky1} F_{z0}\,
    \frac{\sin\!\bigl(2 \arctan(F_{z}/(p_{Ky2}\,F_{z0}))\bigr)}{C_y D_y},\\
    E_y &= p_{Ey1} + p_{Ey2}\,df_z.      & &
\end{align*}

Here, $\alpha$ denotes the axle slip angle, $F_z$ is the corresponding vertical load, and, for each axle, the coefficient vector $\boldsymbol{p}$ collects the Magic Formula parameters that characterize the lateral-force curve and is treated as the set of calibration variables to ensure consistency between the lateral forces of the simulator model and those of the single-track surrogate.

To reconcile the two-axle model with the 14-DoF one, we adopted the following identification workflow:

(i) Data acquisition. For tire calibration, preliminary simulator sessions were run by driving the 14-DoF vehicle model on the same track used for the planning studies, the Siena kart circuit scaled $\times 2$ to better match the dynamic characteristics of the vehicle considered here. These sessions were used to collect the \emph{training data} for tire identification: from the recorded telemetry, axle slip angles were computed as the mean of the left/right tire slip angles,
\(
\bar{\alpha} = \frac{\alpha_{\text{left}} + \alpha_{\text{right}}}{2},
\)
while axle vertical and lateral forces were obtained as
\(
\bar{F}_z = F_z^{\text{left}} + F_z^{\text{right}},\;
\bar{F}_y = F_y^{\text{left}} + F_y^{\text{right}}.
\)

(ii) Parameter identification. Once the training data had been assembled, the single-track tire model was calibrated to reproduce the simulator forces by solving a nonlinear least-squares problem. Denoting by $\boldsymbol{p}$ the Magic Formula parameter vector associated with a given axle, the calibration problem for that axle was posed as
\begin{equation}
		\bar{\boldsymbol{p}} = \arg\min_{\boldsymbol{p}}\; \sum_{k=1}^{N_{\text{train}}} \bigl( \bar{F}_{y,k} - F_{y}(\bar{\alpha}_k,\bar{F}_{z,k};\boldsymbol{p}) \bigr)^2,
	\label{eq:ls_calib}
\end{equation}
where $\bar{F}_{y,k}$ denotes the lateral axle force extracted from the 14-DoF simulator training data at sample $k$, and $F_{y}(\bar{\alpha}_k,\bar{F}_{z,k};\boldsymbol{p})$ is the corresponding force predicted by the single-track tire model, as defined in~\eqref{eq:mf_simplified}, for the same axle slip angle $\bar{\alpha}_k$ and vertical load $\bar{F}_{z,k}$. The optimization~\eqref{eq:ls_calib} is solved independently for the front and rear axles, starting from an initial guess given by the parameter values in the \texttt{.tir} files of the simulator model, yielding distinct parameter vectors for each and thereby aligning the single-track lateral-force response with the 14-DoF telemetry over the training set.

The numerical values of the calibrated Magic Formula parameters collected in $\bar{\boldsymbol{p}}$ for each axle, together with the starting values used to initialize the optimization, are summarized in Table~\ref{tab:tire_par}.

\begin{table}[h]
	\centering
	\caption{Calibrated Magic Formula parameter vectors $\bar{\boldsymbol{p}}$ for the front and rear axles.}
	\begin{tabular}{lcccccccc}
		\hline
		 & $F_{z0}$ (N) & $p_{Cy1}$ & $p_{Dy1}$ & $p_{Dy2}$ & $p_{Ey1}$ & $p_{Ey2}$ & $p_{Ky1}$ & $p_{Ky2}$ \\
		\hline
		Starting Values & $6500$  & $1.45$  & $1.09$  & $-0.20$ & $0.81$ & $0.34$  & $26.94$ & $3.20$ \\
		Front axle & $11050$ & $2.47$ & $1.85$ & $-0.34$ & $0.57$ & $0.59$  & $28.29$ & $3.04$ \\
		Rear axle  & $9210$  & $1.92$  & $1.03$  & $-0.34$ & $0.57$ & $0.59$  & $28.29$ & $3.04$ \\
		\hline
	\end{tabular}
	\label{tab:tire_par}
\end{table}

\begin{figure}[h]
	\centering
	\includegraphics[width=0.8\columnwidth]{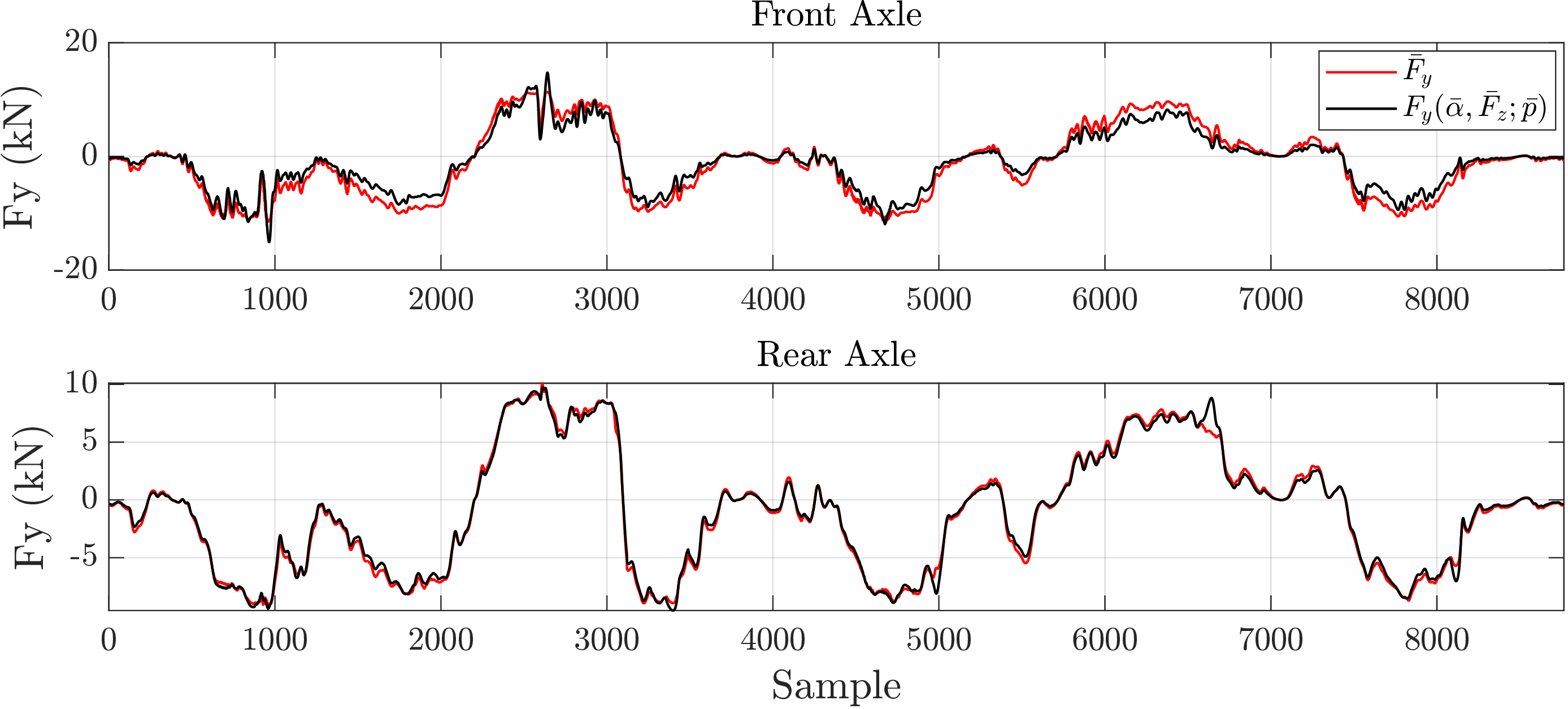}
	\caption{Comparison between axle lateral forces obtained from the simulator telemetry (red) and from the calibrated Magic Formula tire model (black) for the front and rear axles.}
	\label{fig:tire_force_fit}
\end{figure}

Figure~\ref{fig:tire_force_fit} compares the axle lateral forces extracted from the simulator telemetry with those predicted by the calibrated Magic Formula tire model at the front and rear axles, showing close agreement over the operating range of interest.

\begin{figure}[h]
	\centering
	\includegraphics[width=0.9\columnwidth]{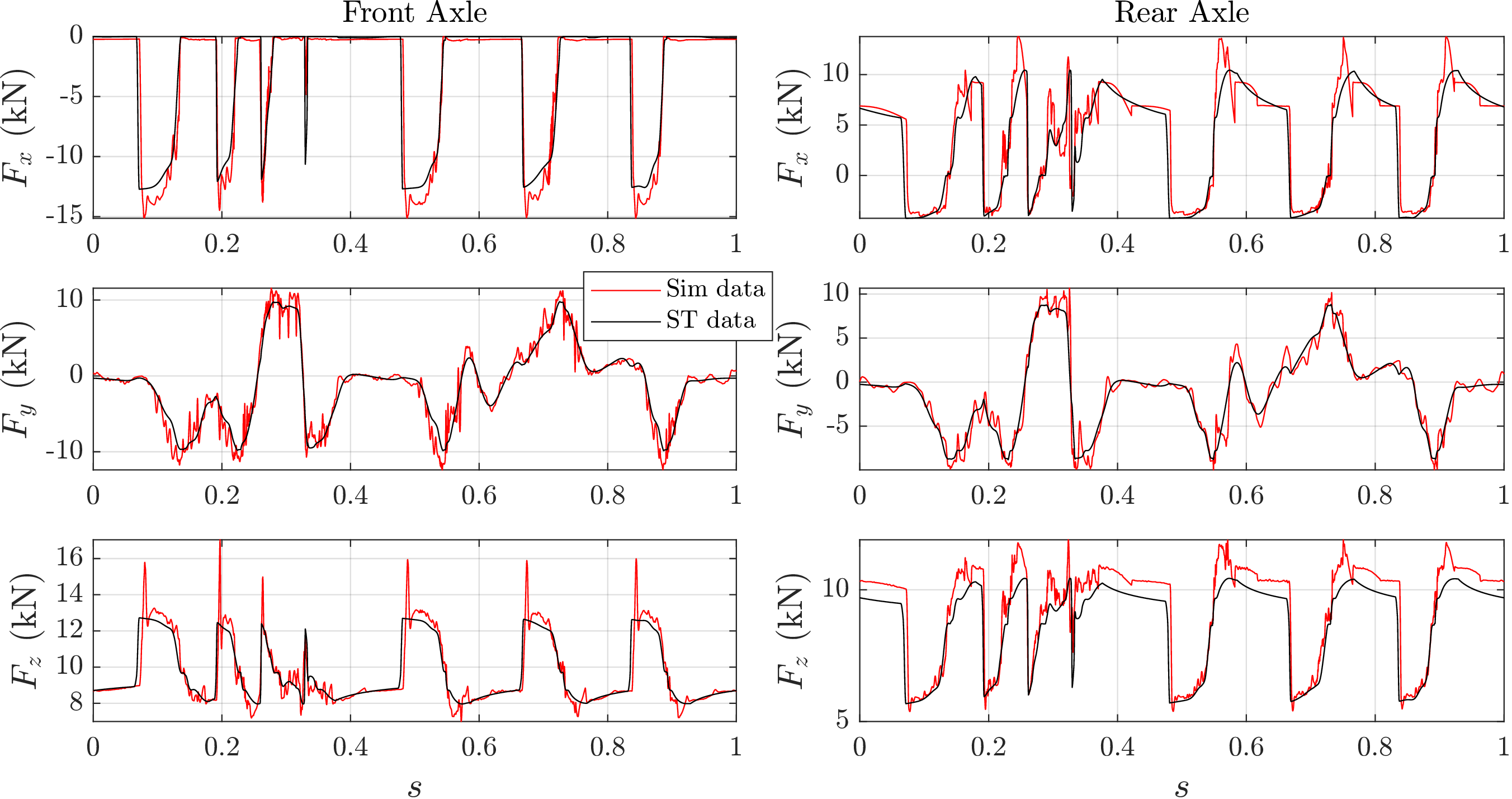}
 	\caption{Comparison between longitudinal, lateral, and vertical axle forces obtained from the simulator (Sim data, red) and from the single-track minimum-lap-time optimization (ST data, black) while attempting to follow the optimized trajectory, plotted against the normalized distances along the track.}
	\label{fig:tire_force_validation}
\end{figure}

(iii) Validation. To validate the calibrated single-track model, we compared data from a Nominal Minimum-lap-time (\NOM) optimization, i.e., without robustness criteria, against simulator data collected with a human driver attempting to track the \NOM \til reference trajectory generated by the lap-time optimization. Figure~\ref{fig:tire_force_validation} illustrates representative comparisons between the longitudinal, lateral, and vertical axle forces obtained from the optimization and those measured in the simulator while the human driver was trying to track the optimized trajectory, confirming the consistency of the single-track model with the high-fidelity simulator dynamics.

\begin{figure}[h]
	\centering
	\includegraphics[width=0.8\columnwidth]{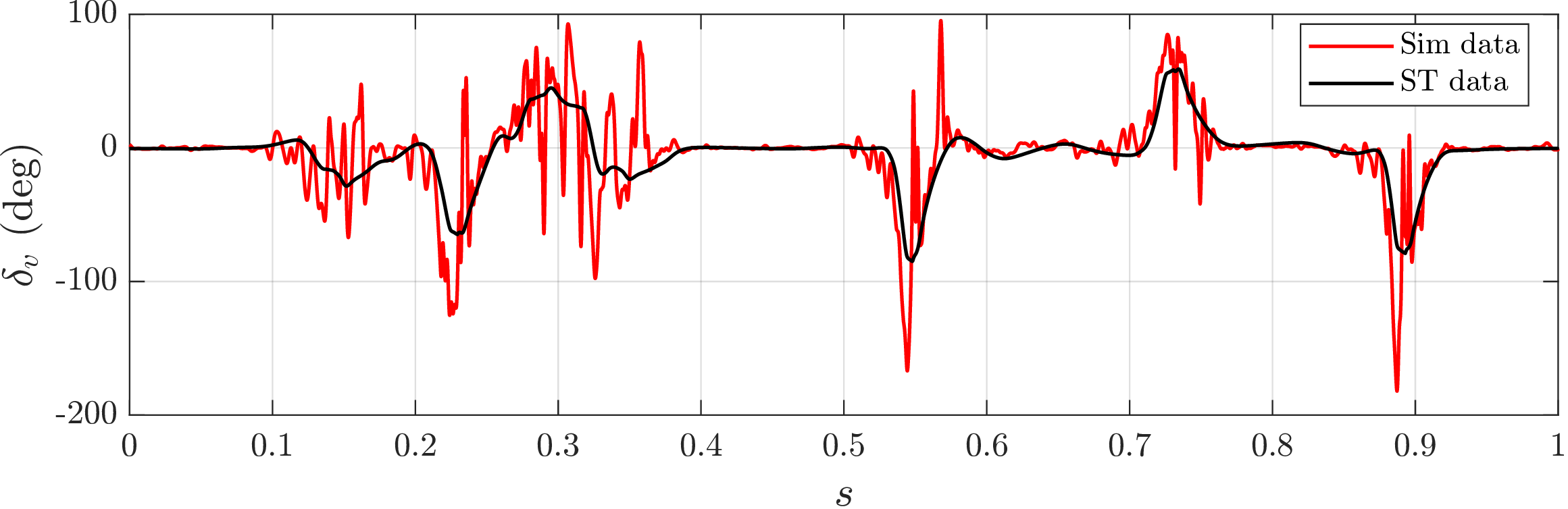}
	\caption{Comparison between steering-wheel angle recorded at the driving simulator (Sim data, red) and steering input from the single-track minimum-lap-time optimization (ST data, black), plotted against the normalized distance along the track while attempting to follow the optimized trajectory.}
	\label{fig:steer_validation}
\end{figure}

Figure~\ref{fig:steer_validation} shows a comparison between the steering-wheel angle recorded at the driving simulator (red) and the steering input generated by the single-track minimum-lap-time optimization (black) along the same reference state trajectory.

\section{Experimental Setting and Test Protocol}
\label{sec:experimental_setting}

\subsection{Drivers, vehicle and track}
Two human drivers participated: \VP{} (Formula SAE and sim-racing experience) and \MM{} (rally background, more expert overall). Both were selected for demonstrated motorsport skills. Trials with passenger-car drivers showed that they could not consistently follow racing-style references: they either deviated from the line or could not sustain the target speed profile, and often incurred catastrophic failures (spins or off-track events requiring a simulator reset), yielding too few valid laps for statistical analysis.

\begin{figure}[h]
	\centering
	\includegraphics[width=0.6\linewidth]{"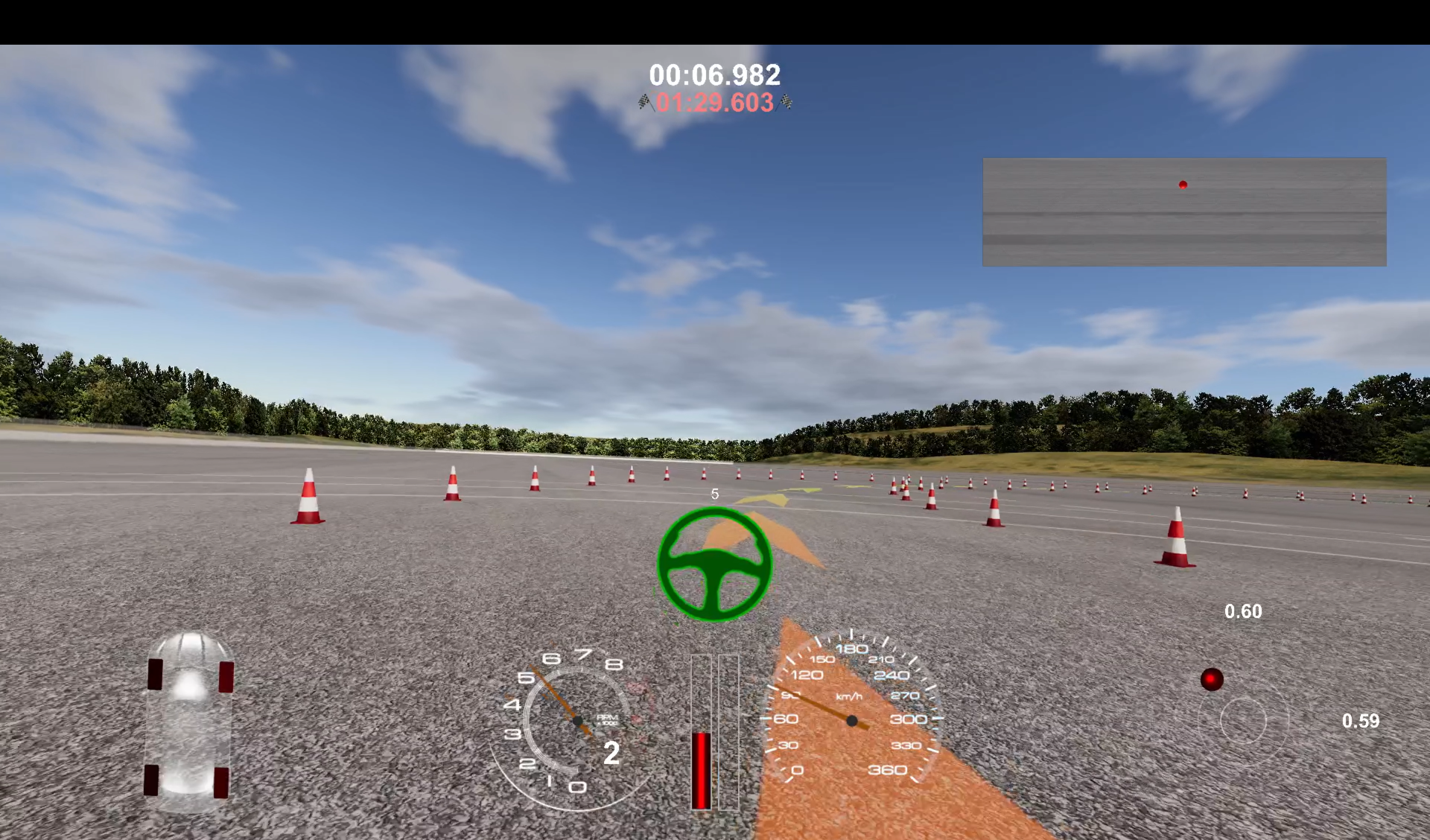"}
	\caption{Dynamic Simulator view from the driver's perspective with the \emph{racing ribbon} overlaid on the scene, indicating the desired trajectory to follow.}
	\label{fig:ribbon_sim}
\end{figure}

The simulator rendered the high-performance car on a closed circuit (the Kartodromo di Siena, scaled $\times 2$ to match the vehicle's dimensions). This track was chosen because it is familiar and offers a balanced mix of fast corners, slow corners, and transitional sections. We conveyed the racing reference as a 1 m-wide \emph{racing ribbon} overlaid on the roadway (Figure \ref{fig:ribbon_sim}). Drivers were instructed to keep their reference point centered within the ribbon while driving. The ribbon's centerline was obtained by translating the planned center-of-mass (CoM) path to the driver reference frame (fixed rigid-body offset), so that the visual cue corresponded to the driver's position rather than to the CoM trajectory. The finite width provided a tolerance band consistent with human execution.

The speed profile was encoded by the ribbon's color map, transitioning from green (full throttle/acceleration) through intermediate hues to red (hard braking). This dual-cue design (lateral ribbon + longitudinal color) offered continuous guidance without numeric overlays or ghost vehicles.

All tests were completed in a single day. For each driver, we randomized and counterbalanced (i) the sequence of trajectory types to be followed (details provided below) and (ii) the order of driver sessions. Rest periods were inserted between blocks. Crucially, the driver was kept unaware of the trajectory category being provided (single-blind), to reduce strategy or effort biases. At the beginning of the day, each driver completed 30 to 40 laps without guidance to build confidence and familiarize themselves with the track layout.

\subsection{Reference trajectories and baseline}
All references were generated with our \emph{lapsim} (optimal-planning) model, well aligned with the simulator dynamics after identification. The track was discretized into 2000 spatial intervals, each 1.3\,m long. We consider four scenarios:
\begin{itemize}[leftmargin=*,nosep]
  \item \NOM: fastest feasible line. The planner minimizes lap time with a small secondary penalty that promotes smooth steering actuation so a human can execute it. The \emph{cost structure is identical} across variants; introducing robustness does not change the planning objective, i.e., lap-time minimization.
  \item \TLC: robust planning with constraint tightening focused on \emph{track limits}, i.e., maintaining explicit margins to the road edges.
  \item \FLC: robust planning with tightening focused on \emph{friction limits}, i.e., maintaining margins to axle/tire saturation and biasing states toward more stable tire utilization.
  \item \NOREF: free driving without a reference, used to assess the impact of following a planned trajectory versus raw driver skill.
\end{itemize}

\begin{figure}
	\centering
	\includegraphics[width=1\linewidth]{"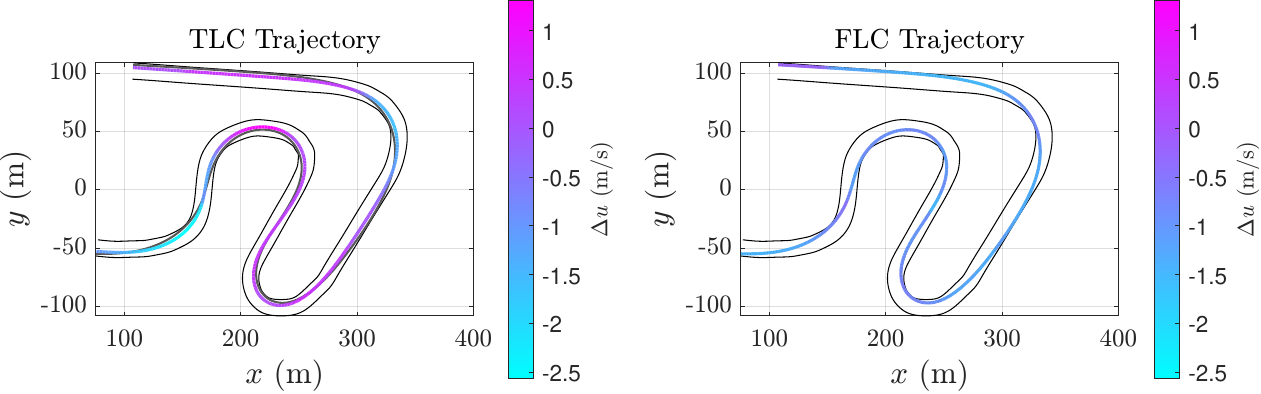"}
	\caption{Comparison of robust planning trajectories (\TLC{} and \FLC{}) overlaid on the nominal trajectory (\NOM{}) for a representative track sector. Trajectory color indicates longitudinal speed difference relative to the nominal. The \TLC{} trajectory (left) deviates more from \NOM{} and achieves higher speeds in some sections, while \FLC{} (right) closely follows \NOM{} at reduced speeds.}
	\label{fig:trajectories_comparison}
\end{figure}

Figure~\ref{fig:trajectories_comparison} shows the two robust trajectories (\TLC{} in the left panel and \FLC{} in the right panel) overlaid on the \NOM{} trajectory in light grey, for a sector of the track. 
The trajectories are colored to highlight the longitudinal speed difference with respect to the nominal. Notably, \TLC{} follows a noticeably different trajectory that allows, in certain portions of the track, a higher speed than \NOM{}, while \FLC{} follows a trajectory that is closer to \NOM{} with a lower longitudinal speed. 

Robust back-offs (\TLC{}, \FLC{}) are computed by resetting the state covariance at each curvilinear grid point and propagating it over a horizon of $H$ steps. In our experiments, setting $H\!=\!4$ results in a maximum delay of 5.2\,m in the driver's corrective actions.
The grid spacing and the reset covariance assumptions were calibrated from pilot tests so that the induced tightenings correspond to a $99\%$ constraint-satisfaction confidence, implemented via a safety factor $\gamma\!=\!3$ (three-sigma). In practice, these settings yield a tangible increase in constraint margins while preserving executability.

\subsection{Metrics}
We evaluate performance and tracking using the following quantities (with respect to the \emph{active} reference, i.e., \NOM, \TLC, or \FLC):

\noindent\textbf{Lap Time (s).} Total lap time LT, in seconds, over valid laps.

\noindent\textbf{Steer energy ($\boldsymbol{\mathrm{rad}^2/\mathrm{s}}$).} As a proxy for \emph{steering activity}, we use
\begin{align}
E_s=\int_0^T \dot{\delta}(t)^2\,dt,
\end{align}
 so lower values indicate smoother, more executable actuation.
We introduce the root-mean-square (RMS) over a time horizon $[0,T]$. For a signal $f(\cdot)$:
\begin{align}
\RMS{f} \;\triangleq\; \sqrt{\frac{1}{T}\int_{0}^{T} f(t)^2\,dt}.
\end{align}
and use it to define both \emph{tracking fidelity} and \emph{style indicators}.

\noindent\textbf{Tracking fidelity.} RMS lateral position and speed errors, measuring adherence to the planned line and speed profile:
\begin{align}
\RMS{e_y}\ \text{(m)},\qquad \RMS{e_v}\ \text{(m/s)}.
\end{align}

\noindent Statistics are computed over all valid laps per driver and condition.


\noindent\textbf{Style indicators.} RMS side-slip usage:
\[
\mathrm{RMS}\,\beta_{\mathrm{drv}}\quad\text{and}\quad \mathrm{RMS}\,\beta_{\mathrm{ref}},
\]
to compare the driver's typical sideslip $\beta_{\mathrm{drv}}$ with that implied by the planned trajectory $\beta_{\mathrm{ref}}$. 

\section{Results} 
\label{sec:results} 
\begin{figure}[h]
  \centering
  \includegraphics[width=0.80\linewidth]{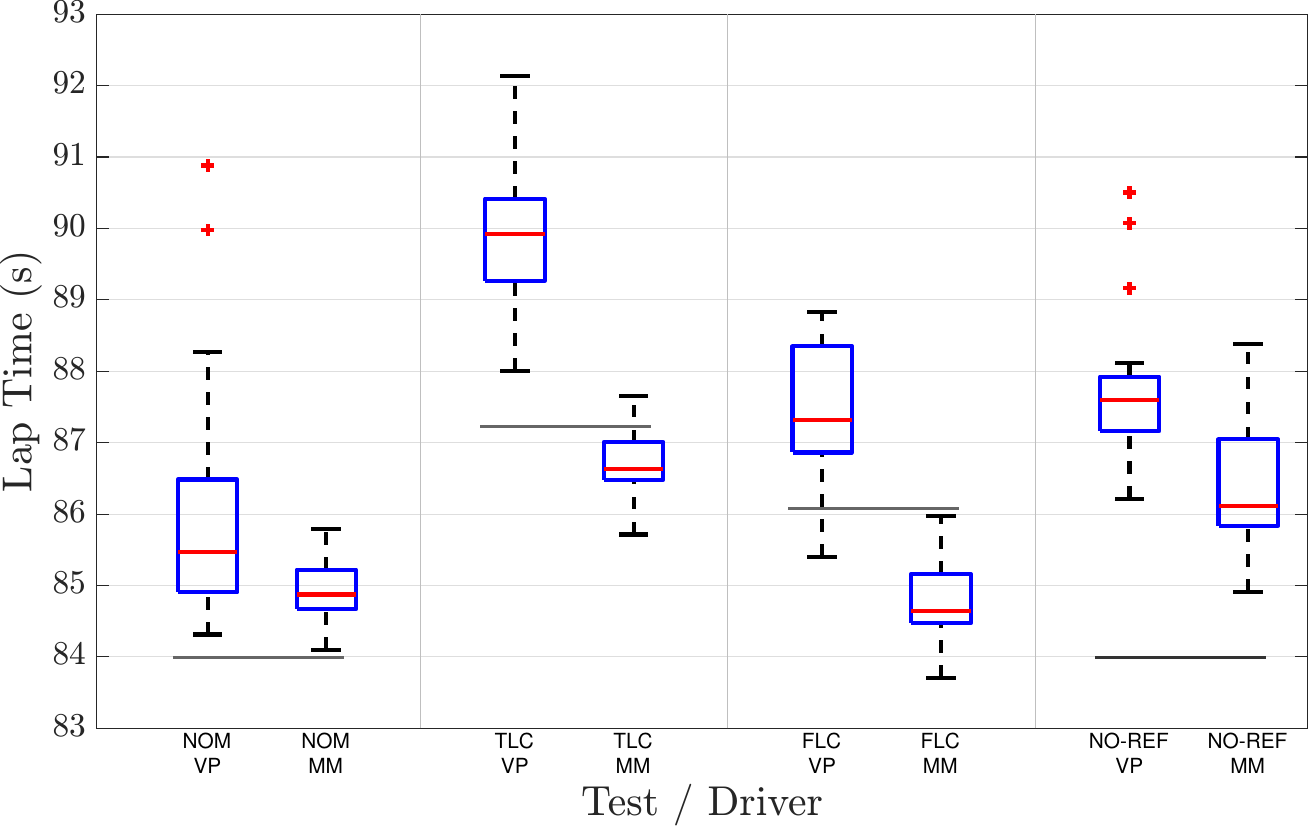}
  \caption{Lap times by test and driver (box plots). Grey lines indicate the reference laptime for each trajectory type. For driver \VP{}, \NOM{} is fastest and \FLC{} is moderately slower; for driver \MM{}, \NOM{} and \FLC{} yield comparable medians; \TLC{} is slower for both; \NOREF{} for driver \MM{} is slower and more variable while driver \VP{} shows a more consistent performance with higher medians.}
  \label{fig:laptime}
\end{figure}
\subsection{Lap-time distributions}
Figure~\ref{fig:laptime} juxtaposes the lap-time (LT) box plots for each trajectory--driver pair and provides the main visual reference for this subsection.

For \VP{}, the leftmost boxes (\NOM{}) anchor the fastest medians, confirming the benchmark role of the nominal planner for this driver.
For \MM{}, however, the \FLC{} and \NOM{} boxes are practically superposed: their medians lie within each other's IQR and any LT gap is visually negligible, indicating comparable lap times under these two references.
Notably, for the robust trajectories (\FLC{} and \TLC{}), driver \MM{} achieves laptimes faster than the planner's reference (indicated by the grey lines), suggesting that the calibration phase can be improved: \MM{} operates with lower margins than those assumed by the planner, effectively exploiting the vehicle envelope more aggressively. This observation connects to the RMS tracking errors discussed in Section~\ref{sec:rms}: \MM{}'s lower trajectory-following fidelity (higher RMS errors, as visible in Figure~\ref{fig:rms}) enables these faster laptimes by deviating from the planned path to exploit tighter margins.
The \FLC{} column for \VP{} shifts vertically by roughly $1$--$2$\,s relative to \NOM{} yet preserves a compact Interquantile Range (IQR), signalling a modest time penalty in exchange for tighter dispersion.
\TLC{} occupies the clearly slower columns in Figure~\ref{fig:laptime}: both drivers show medians several seconds above \NOM{}, consistent with the more conservative curvature shaping.
Finally, the \NOREF{} column for \MM{} exhibits the tallest whiskers and a median higher than \NOM{} and \FLC{}, highlighting the variability that accompanies unguided laps.
For \NOREF{} runs, both drivers show medians above their \NOM{} and \FLC{} benchmarks, but with distinct dispersion patterns: \VP{} displays relatively short whiskers with a few isolated outliers, whereas \MM{} shows taller whiskers without clear outliers, indicating more run-to-run variability in typical laps but no extreme excursions.
\begin{figure}[t]
  \centering
  \includegraphics[width=0.80\linewidth]{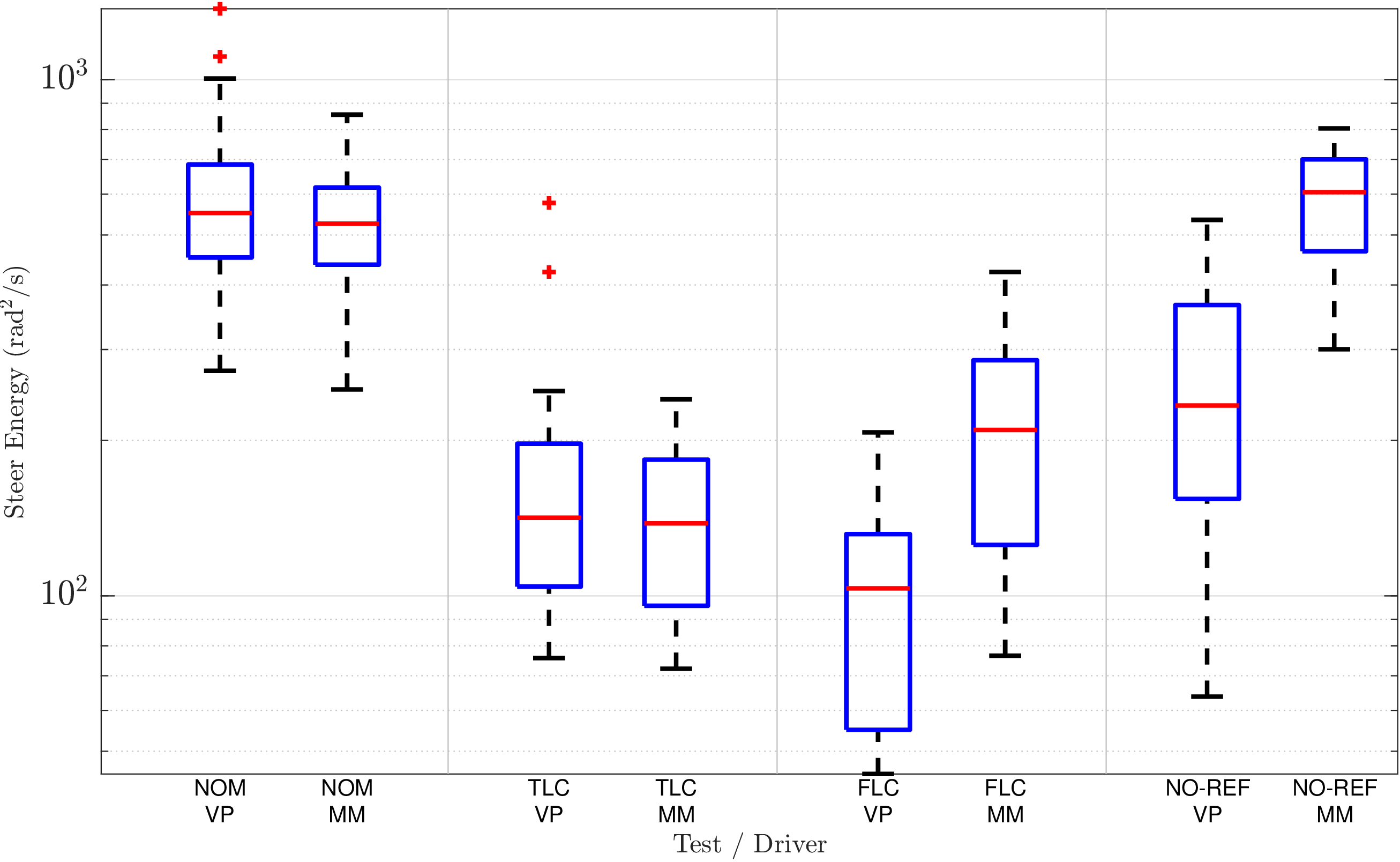}
  \caption{Steer Energy by test and driver (log scale). \NOM{} requires the highest effort among reference-following tests; \FLC{} and \TLC{} markedly reduce effort. For driver \MM{}, \NOREF{} shows \emph{high} SE coupled with slow LTs.}
  \label{fig:se}
\end{figure}

\subsection{Steering effort}
Figure~\ref{fig:se} mirrors the structure of Figure~\ref{fig:laptime} but maps the steer energy SE distributions on a logarithmic axis, making the effort differences visually salient.
The \NOM{} boxes and the \NOREF{} box for \MM{} occupy the upper part of the axis, meaning that both drivers pay the highest steering cost whenever the nominal reference is enforced or free driving is performed.
Moving to the \FLC{} and \TLC{} columns in the same figure reveals remarkable effort relief: for \VP{}, \FLC{} reaches the lowest median SE, while \MM{} exhibits the ordering \(\TLC < \FLC < \NOM\) with partial overlap between \FLC{} and \TLC{} quartiles.

The rightmost column (\NOREF{} for \MM{}) should be read together with the LT information: its elevated median and wide spread in Figure~\ref{fig:se}, combined with the slow and dispersed LT box in Figure~\ref{fig:laptime}, illustrate that free driving simultaneously degrades time and effort metrics relative to the reference-guided alternatives, especially for \NOM{} and \FLC{}.

\subsection{Time vs effort Pareto structure}
\begin{figure}[t]
  \centering
  \includegraphics[width=0.85\linewidth]{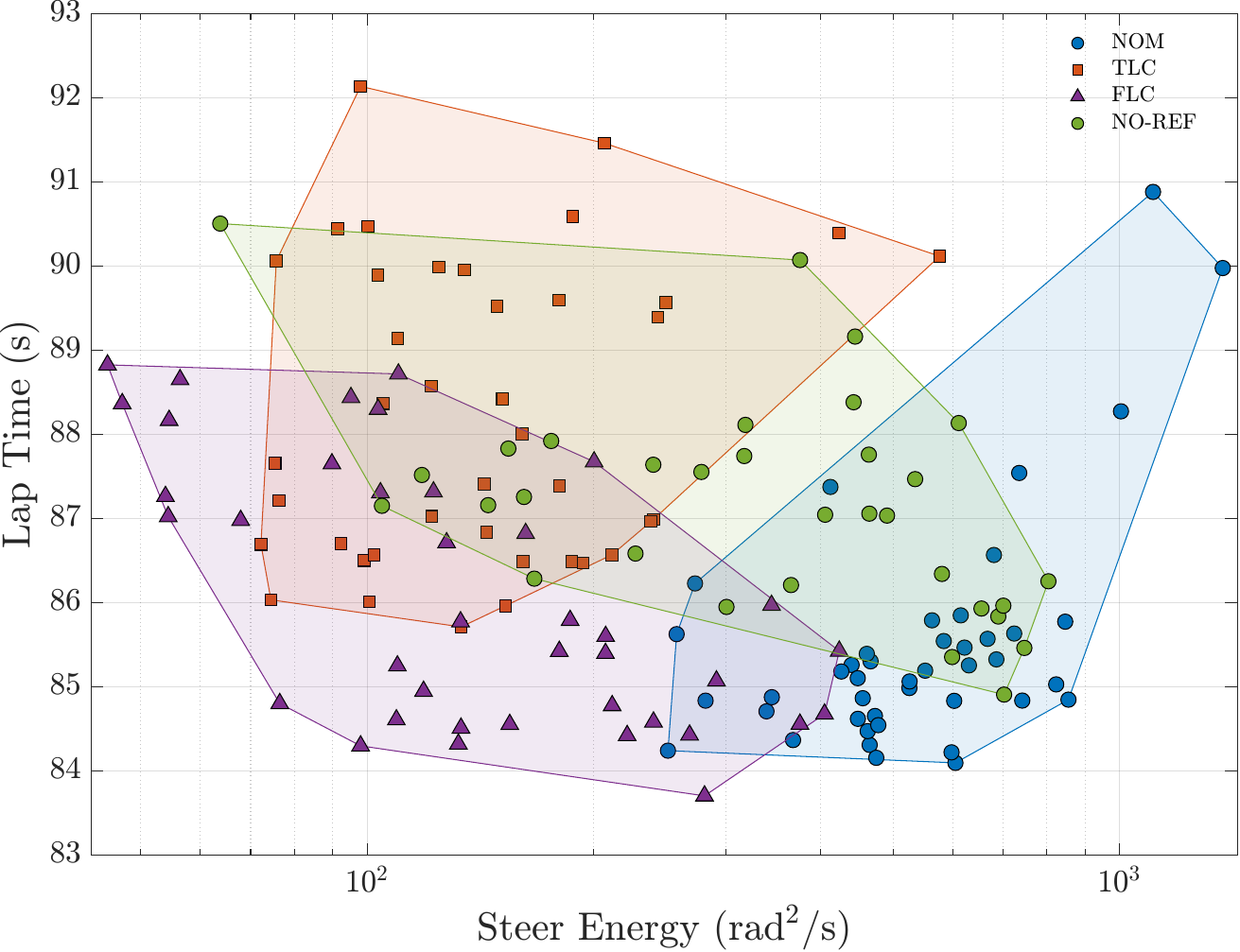}
  \caption{Overall Lap Time--Steer Energy scatter plot (test types as colors, both drivers). \FLC{} typically lies near a Pareto-efficient band; \NOM{} is \emph{generally}\protect\footnotemark\hspace{0.5ex} fastest but costly; TLC is lowest effort but slower.}
  \label{fig:overall-scatter}
\end{figure}

\begin{figure}[t]
  \centering
  \includegraphics[width=0.85\linewidth]{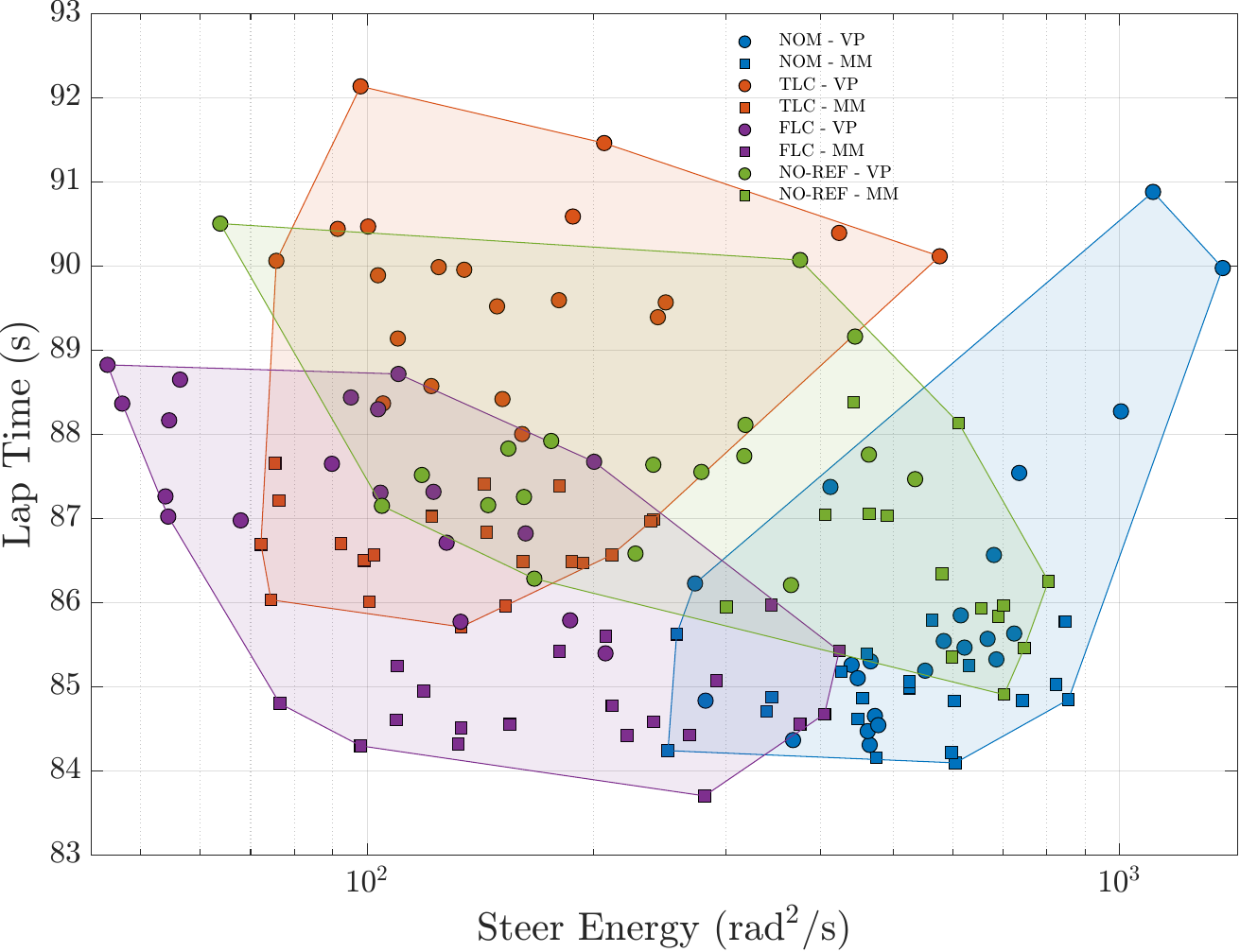}
  \caption{Lap Time--Steer Energy scatter plot with driver markers (colors = test types; markers distinguish driver \VP{} and driver \MM{}). \FLC{} retains the favorable band for both drivers; \NOREF{} for driver \MM{} sits at high SE and high LT.}
  \label{fig:comp-scatter}
\end{figure}

Figures~\ref{fig:overall-scatter} and~\ref{fig:comp-scatter} condense the box-plot insights into two scatter plots that directly show the LT--SE trade-off.
In Figure~\ref{fig:overall-scatter}, markers colored by the reference-guided trajectories fall into three recognizable clusters: \NOM{} points accumulate along the lower boundary in LT but sit high in SE, \TLC{} marks gravitate toward the upper-left (slow but economical), and \FLC{} forms an elongated ribbon close to an empirical lower-left Pareto front linking intermediate LTs with markedly lower SE.
Separating the runs by driver in Figure~\ref{fig:comp-scatter} clarifies how this ribbon behaves: the boxed markers for \MM{} lie near the fast end of the \FLC{} band, whereas the circle markers for \VP{} extend the same band toward slightly longer LTs while maintaining a considerable effort gap to \NOM{}.
As shown in Figures~\ref{fig:overall-scatter} and~\ref{fig:comp-scatter}, the \NOREF{} trajectory confirms the picture anticipated in the previous section: \VP{} exhibits a smaller dispersion in lap time at a moderate steering-energy level, with his runs clustering near the middle of the scatter plot, whereas \MM{} populates the right-hand side of the plot at higher steering-energy levels with a much larger spread in lap times.

\footnotetext{This is because only the lowest lap-time sample corresponds to \TLC.}

\subsection{RMS tracking errors and activity}
\label{sec:rms}
\begin{figure}[t]
  \centering
  \includegraphics[width=0.85\linewidth]{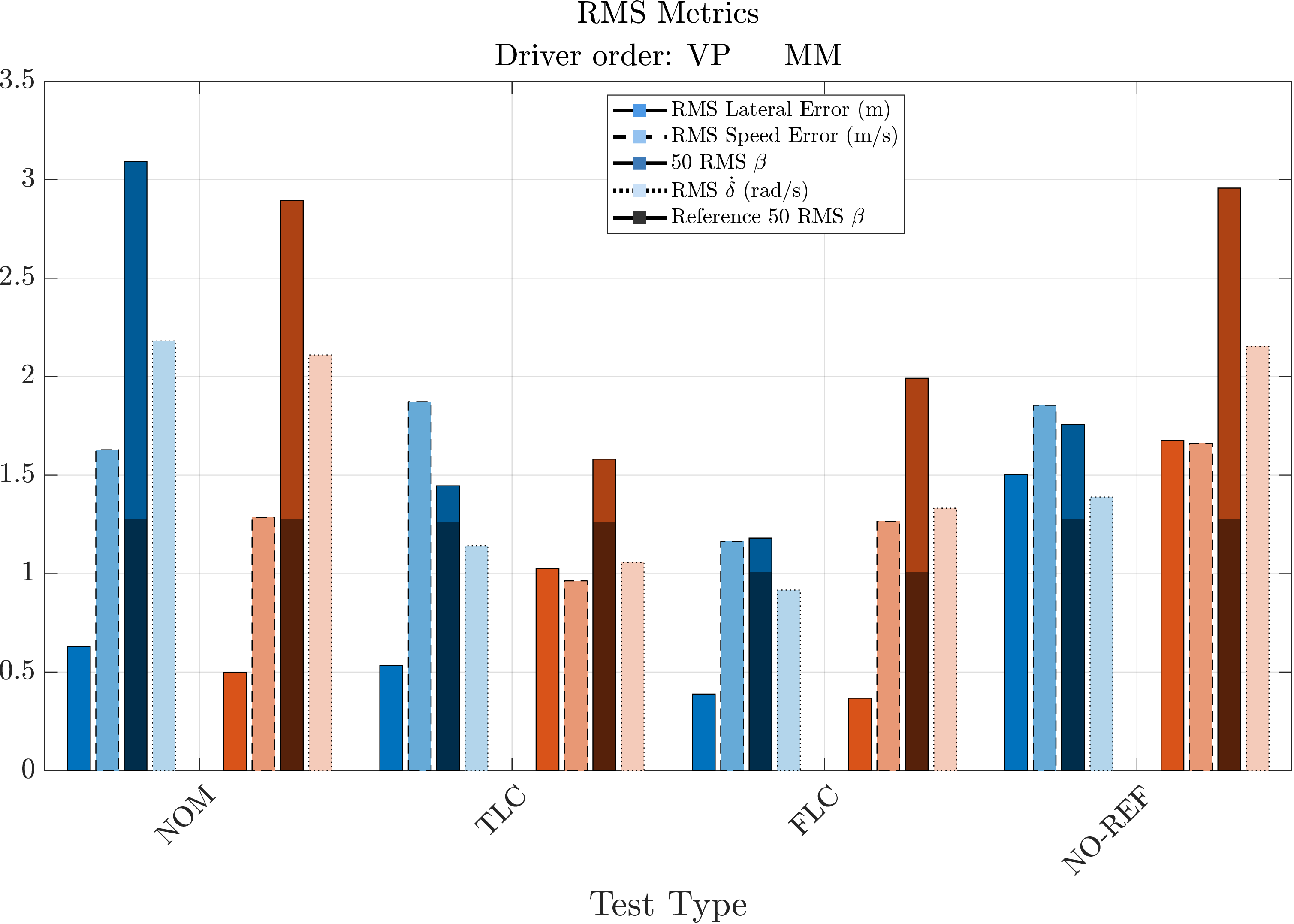}
  \caption{RMS metrics by test and driver (driver order: \VP{} then \MM{}). \FLC{} eases tracking vs.\ \NOM{} and reduces steering activity; \TLC{} is most ``relaxed'' but slowest; \NOREF{} for \MM{} combines poor tracking with elevated SE and slow LTs.}
  \label{fig:rms}
\end{figure}

Figure~\ref{fig:rms} reports, for each reference condition, two four-bar clusters---left for \VP{} and right for \MM{}. Within each cluster, four quantities are shown: (i) $\RMS{e_y}$, (ii) $\RMS{e_v}$, (iii) $\RMS{\beta_{\text{drv}}}$ with the corresponding $\RMS{\beta_{\mathrm{ref}}}$ overlaid in black, and (iv) a light-colored bar for $\RMS{\dot{\delta}}$ as a qualitative proxy for SE.

Across conditions, \FLC{} typically pulls $\RMS{e_y}$ and $\RMS{e_v}$ below \NOM{}, while simultaneously reducing $\RMS{\dot{\delta}}$; the effect is clearest for \VP{} but remains visible for \MM{}. Under \TLC{} both drivers see reduced $\RMS{\beta_{\text{drv}}}$ and $\RMS{\dot{\delta}}$---consistent with its more relaxed trajectory shaping---but at the price of increased \LT{} and tracking errors: \VP{} experiences a noticeable rise in speed error $\RMS{e_v}$, whereas \MM{} shows larger lateral error $\RMS{e_y}$ relative to \NOM{}. According to the drivers' qualitative feedback, this is because the \TLC{} path feels somewhat unnatural to follow, tending to keep close to the lane center rather than supporting the corner-cutting strategies they would normally adopt. 
The larger RMS errors observed for \MM{} under both \FLC{} and \TLC{} correlate with his faster-than-planner laptimes (visible in Figure~\ref{fig:laptime}): by deviating more from the planned trajectory than \VP{}, \MM{} exploits lower margins and achieves superior performance, indicating that the planner's conservative calibration leaves room for skilled drivers to operate closer to the vehicle limits. 

Beyond these aggregate trends, the \NOREF{} cluster also highlights differences in driving style and background between the two drivers: \VP{}, who is less experienced and more conservative, exhibits lower $\RMS{\beta_{\text{drv}}}$ and $\RMS{\dot{\delta}}$ than in the \NOM{} condition, where he attempts to push harder, whereas \MM{}, the more experienced driver, operates closer to the vehicle limits and displays a rally-style behavior with larger $\RMS{\beta_{\text{drv}}}$ and $\RMS{\dot{\delta}}$ under \NOREF{} than when following reference trajectories. This further indicates that, when a reference is provided, both drivers deliberately move away from their natural style. Crucially, the \emph{natural} behavior observed in \NOREF{} is not a \emph{comfort zone}, especially for \MM{}: it yields worse performance and higher energy expenditure, because the free-run trajectories they gravitate toward are, paradoxically, less stable and demand more corrective action--hence a larger steering effort SE.

\subsection{Summary tables (medians and IQR)}
To complement the description above with concise descriptors, in Tables~\ref{tab:lt} and~\ref{tab:se} we report \emph{medians} and \emph{interquartile ranges} (IQR) that are extracted from Figures~\ref{fig:laptime} and~\ref{fig:se}. 
The median captures the central lap-time or steer-energy tendency even when distributions are skewed, while the IQR (the span between the 25th and 75th percentiles) reflects how tightly each controller concentrates its performance.
Because our sample sizes per condition are modest, the IQR is especially informative: a narrow band signals repeatable behavior, whereas a wide one warns that the observed median may be sensitive to session-to-session variability.

Taken together, Tables~\ref{tab:lt}--\ref{tab:deltas} sharpen the picture sketched by the box plots. For \VP{}, \FLC{} trades a modest $2\%$ increase in LT for an $81\%$ reduction in SE relative to \NOM{}, while \TLC{} pushes LT up by a further $3\%$ with only a slight SE gain over \FLC{}. For \MM{}, \FLC{} essentially preserves LT (a $0.3\%$ improvement vs.\ \NOM{}) while still cutting SE by about $60\%$, whereas \TLC{} delivers an additional $34\%$ SE reduction at the cost of a $2\%$ LT increase. The \NOREF{} rows confirm that free driving is systematically dominated: both drivers incur longer LTs than under \NOM{}/\FLC{} and substantially higher SE, with \MM{} in particular combining the slowest and most variable times with the largest effort.

\begin{table}[h!]
\centering
\caption{Lap Time (s): medians and IQR by driver and test.}
\begin{tabular}{lcccc}
\toprule
 & \NOM & \TLC & \FLC & \NOREF \\
\midrule
\VP{} Median [IQR] & $85.47$ [$1.58$] & $89.92$ [$1.15$] & $87.32$ [$1.49$] & $87.60$ [$0.76$] \\
\MM{} Median [IQR] & $84.87$ [$0.55$] & $86.63$ [$0.53$] & $84.65$ [$0.69$] & $86.11$ [$1.21$] \\
\bottomrule
\end{tabular}
\label{tab:lt}
\end{table}

\begin{table}[h!]
\centering
\caption{Steer Energy (rad$^2$/s): medians and IQR by driver and test (log-scale plots).}
\begin{tabular}{lcccc}
\toprule
 & \NOM & \TLC & \FLC & \NOREF \\
\midrule
\VP{} Median [IQR] & $552$ [$232$] & $142$ [$93$] & $103$ \phantom{0}[$77$] & $234$ [$212$] \\
\MM{} Median [IQR] & $526$ [$180$] & $138$ [$88$] & $210$ [$161$] & $605$ [$236$] \\
\bottomrule
\end{tabular}
\label{tab:se}
\end{table}

\begin{table}[h!]
\centering
\caption{Pairwise median differences (FLC vs NOM; TLC vs FLC). Negative $\Delta$ in SE denotes effort reduction.}
\begin{tabular}{lcccc}
\toprule
 & \multicolumn{2}{c}{\VP{}} & \multicolumn{2}{c}{\MM{}} \\
\cmidrule(lr){2-3}\cmidrule(lr){4-5}
 & $\Delta$LT [s] & $\Delta$SE [rad$^2$/s] & $\Delta$LT [s] & $\Delta$SE [rad$^2$/s] \\
\midrule
FLC $-$ NOM & $+1.85$ ($+2\%$) & $-449$ ($-81\%$) & $-0.22$ ($-0.3\%$) & $-316$ ($-60\%$) \\
TLC $-$ FLC & $+2.60$ ($+3\%$) & \phantom{0}$+39$ ($+38\%$) & $+1.98$ \phantom{0.}($+2\%$) & \phantom{0}$-72$ ($-34\%$) \\
\bottomrule
\end{tabular}
\label{tab:deltas}
\end{table}

\subsection{Discussion}
\paragraph{Finding 1 (drivability trade-off, preliminary).}
Taken together, Figures~\ref{fig:laptime}, \ref{fig:se}, and~\ref{fig:rms} position \FLC{} on a ``reduced-effort / comparable-time'' band: for driver \MM{} the LT gap to \NOM{} is within the quartiles of the benchmark box plot, whereas for driver \VP{} the penalty reaches $1$--$2$\,s yet coincides with the lowest SE box in Figure~\ref{fig:se} and a marked drop in RMS indicators.
Given the modest number of laps per condition, this constellation should be read as an encouraging trend rather than a definitive performance guarantee.

\paragraph{Finding 2 (conservatism lever).}
The combination of Figures~\ref{fig:laptime} and~\ref{fig:se} indicates that \TLC{} achieves low SE medians (the lowest among all conditions for driver \MM{}), but at the cost of a marked increase in LT. In motorsport, where lap time is the primary objective, this effort-time trade-off is generally unacceptable. Moreover, \TLC{}'s margin-keeping geometry discourages decisive corner cutting, lengthens paths, and depresses apex speeds--yielding trajectories that feel less natural to fast drivers despite the reduced effort.


\paragraph{Finding 3 (driver modulation).}
Figures~\ref{fig:comp-scatter} and~\ref{fig:laptime} jointly indicate that the LT penalty under \FLC{} depends on the driver: \MM{} keeps lap times close to \NOM{} (clustered at the fast end of the \FLC{} ribbon), whereas driver \VP{} trades about $1$--$2$\,s for the same controller.
More strikingly, \MM{} exceeds the planner's reference laptime for both \FLC{} and \TLC{} (as indicated by the grey lines in Figure~\ref{fig:laptime}), demonstrating that skilled drivers can operate with lower margins than those assumed during the planner's calibration phase. This performance advantage comes at the cost of trajectory fidelity: \MM{}'s higher RMS tracking errors (Figure~\ref{fig:rms}) reflect his deliberate deviations from the planned path to exploit tighter margins, revealing a trade-off between strict trajectory following and laptime performance that the planner's conservative calibration does not capture.
Free driving of \MM{} (\NOREF{}) reinforces this driver effect: in both figures the corresponding points occupy the high-SE, medium-high-LT side, suggesting that the reference simplifies the exploitation of the vehicle envelope.

\paragraph{Implications and open questions.}
The qualitative structure in Figures~\ref{fig:laptime}--\ref{fig:rms} hints that \FLC{} biases closed-loop operation away from abrupt friction-utilization changes, whereas \NOM{} remains sensitive to modeling mismatch and \TLC{} slows the car by maintaining distance from geometric limits.

Future experiments should probe whether these patterns persist with more participants, alternative tracks, longer exposure, and richer effort proxies (including cognitive-load measures), and should explicitly quantify sequencing effects that are not controlled in the present dataset.

\section{Conclusion}
\label{sec:conclusion}
The experiments demonstrate that disturbance-aware planned trajectories yield
distinct operating regimes. We recall that \NOM{} denotes the nominal
time-optimal trajectory, \TLC{} a track-limit-robust time-optimal trajectory
obtained by tightening margins to the track edges, and \FLC{} a
friction-limit-robust time-optimal trajectory obtained by tightening against
axle/tire saturation. Apart from a single sample, \NOM{} retains the lap-time
benchmark overall but imposes the highest steer-energy expenditure among the
reference-following conditions; \TLC{} markedly lowers effort yet elongates
trajectories, eroding its usefulness in lap-time-driven contexts; and \FLC{}
remains close to the empirical LT--SE Pareto band, delivering substantial
effort savings with only marginal penalties, especially for driver \MM{}.
Notably, driver \MM{} achieved laptimes faster than the planner's reference for both \FLC{} and \TLC{}, indicating that the planner's calibration margins may be conservative for skilled drivers who can operate closer to vehicle limits. This performance advantage correlates with \MM{}'s higher RMS tracking errors, reflecting his deliberate deviations from the planned trajectory to exploit tighter margins---a finding that suggests the calibration phase can be refined to better match driver capabilities.
Free driving (\NOREF{}) offers no significant advantage: medians for \MM{}
reach $86.11$\,s and $605$\,rad$^2$/s, confirming that unguided laps are
simultaneously slower and costlier.

These observations position \FLC{} as the most promising compromise for
applications that must balance pace and driver workload, including endurance
stints, acclimation to unfamiliar circuits, or vehicle configurations with
pronounced uncertainty.

The study remains exploratory---two drivers and limited laps per condition---so the findings should be validated with broader participant pools, additional tracks, and richer workload indicators (cognitive or physiological). Such extensions will clarify how persistent the favorable \FLC{} band and the dominated \NOREF{} regime are when exposure time, learning effects, and driver diversity increase. 





\section*{Funding}
This work is supported by project PRIN 2022 PNRR ``Global Stability of road vehicle
motion - STAVE'' CUP I53D23005670001.


\bibliography{references}

\end{document}